\documentclass{article} % For LaTeX2e
\usepackage{nips15submit_e,times}
\usepackage{hyperref}
\usepackage{url}
\usepackage{authblk}
\usepackage{amsmath,graphicx,booktabs}
\usepackage{epstopdf,amssymb}
\usepackage{multirow}
\usepackage{fixltx2e}
\usepackage{subfigure}
\usepackage{float}
\usepackage{titlesec}
\usepackage{bm}
\usepackage{mathrsfs}

\title{\textbf{Explanation of CNN Mechanism Using Traditional Facial Affective Recognition}} %Autocontex
\author{Yongpei Zhu$^{1}$, Hongwei Fan$^{1}$, Kehong Yuan$^{1*}$\\
	$^{1}$Graduate School at Shenzhen, Tsinghua University, Shenzhen 518055, China.\\
	*Corresponding author: Kehong Yuan (e-mail: yuankh@sz.tsinghua.edu.cn)\\
	\texttt{zhuyp17@mails.tsinghua.edu.cn}\\
}

\nipsfinalcopy % Uncomment for camera-ready version

\begin{document}
	
	\maketitle
	
	\begin{abstract}
		With the development of deep learning, the structure of convolution neural network is becoming more and more complex and the performance of object recognition is getting better. However, the classification mechanism of convolution neural networks is still an unsolved core problem. The main problem is that convolution neural networks have too many parameters, which makes it difficult to analyze them. In this paper, we design and train a convolution neural network based on the expression recognition, and explore the classification mechanism of the network. By using the Deconvolution visualization method, the extremum point of the convolution neural network is projected back to the pixel space of the original image, and we qualitatively verify that the trained expression recognition convolution neural network forms a detector for the specific facial action unit. At the same time, we design the distance function to measure the distance between the presence of facial feature unit and the maximal value of the response on the feature map of convolution neural network. The greater the distance, the more sensitive the feature map is to the facial feature unit. By comparing the maximum distance of all facial feature elements in the feature graph, the mapping relationship between facial feature element and convolution neural network feature map is determined. Therefore, we have verified that the convolution neural network has formed a detector for the facial Action unit in the training process to realize the expression recognition.\\
		
	\end{abstract}
	
%	\begin{abstract}
%		
%	\end{abstract}
%	
%	\begin{keyword}
%		\MSC 41A05\sep 41A10\sep 65D05\sep 65D17
%		\KWD deconvolution\sep CNN\sep expression recognition
%		
%		%% MSC codes here, in the form: \MSC code \sep code
%		%% or \MSC[2008] code \sep code (2000 is the default)
%	\end{keyword}
%	
%\end{frontmatter}

%\linenumbers

%% main text
\section{Introduction}
\label{sec1}
Deep learning technology is profoundly changing the course of human history. It has brought earth-shaking revolution in almost all fields of information science, such as image recognition, speech recognition, natural language processing and text translation.
The great success of deep learning is mainly attributed to the convolutional neural network.
The purpose of this paper is to provide a reasonable explanation for the effectiveness of convolutional neural networks in deep learning.So what natural law is responsible for the great success of deep learning?The success of deep learning is based on the inherent laws of the data itself, which deep learning technology can reveal and utilize.High-dimensional data is distributed near low-dimensional manifolds with specific probability distribution on manifolds, and the strong ability of deep learning network to approach nonlinear mapping.
Deep learning technology can extract the manifold structure from a class of data and express the overall prior knowledge in the form of a manifold, specifically, the encoding and decoding mapping, which is implied in the weight of the neurons in CNN.

In recent years, the use of convolution neural network in deep learning\cite{lecun2015deep} for feature extraction and classification has become more and more mature and popular. The use of convolution neural network for feature extraction has higher accuracy and stronger robustness. As more and more sophisticated network structures are designed, many image classification problems in computer vision have been well solved, such as face recognition and object detection in some open data sets have been greatly improved the accuracy.

However, in many cases convolution neural network is used as a black box. The mechanism of internal classification is not clear. Most of the time only through experience and constantly trying to improve the network, which is not conducive to the improvement of the neural network.

Several groups began to look into this black-box problem, that review is by Davide Castelvecchi in 2016\cite{castelvecchi2016can}.  The team of  Geoffrey Hinton began to look into this black-box problem from 2012 and to dig deeper into how this was possible,  Vedaldi¡¯s group took algorithms that Hinton had developed to improve neural-network training, and essentially ran them in reverse. Rather than teaching a network to give the correct interpretation of an image, the team started with pretrained networks and tried to reconstruct the images that produced them. This helped the researchers to identify how the machine was representing various features??as if they were asking a hypothetical cancer-spotting neural network Clune's team discovered in 2014 that the black-box problem might be worse than expected: neural networks are surprisingly easy to fool with images that to people look like random noise, or abstract geometric patterns.

Therefore, this paper has done the following work: design a network for expression classification, try to analyze and verify the principle of the network classification. Finally, there comes to a conclusion that the feature which the neural network get from learning is some facial action units and it is very similar to the Facial Action Coding System. The reason to choose facial expression recognition is because it is a representative classification problem and a lot of study has been done. There is a large number of academic data sets available.The innovation of this paper is to explain the learning mechanism of deep learning, exploring the correspondence between facial feature units and feature maps with deconvolution visualization, and explain the learning mechanism of CNN with the help of expression classification.

\section{Manifold learning in deep learning}
According to Gu's work,we can use manifold geometry to study the essence of deep learning.The fundamental laws (or assumptions) of deep learning can be concluded as follows:

1. Manifold Distribution Law:High-dimensional data of one category in nature tend to be concentrated around a low-dimensional manifold.

2. Clustering Distribution Law:Different subclasses in one category correspond to different probability distributions on manifolds, and the distance between these distributions is large enough to distinguish these subclasses.

One of the main purposes and functions of deep learning is to learn hidden manifolds and probability distributions on manifolds from data.
\subsection{Manifold structure}
Manifold is the most basic concept in topology and differential geometry, which is essentially a space composed of many Euclidean Spaces pasted together.As shown in Figure 1, A manifold (composed of two-dimensional face images) is a topology space $\mathrm{S}$, which is covered by a family of open sets $\mathrm{S}\subset{{\cup}_{\alpha}{\mathrm{U}_{\alpha}}}$, and there is a homeomorphic mapping ${\varphi}_{\alpha}:{\mathrm{U}_{\alpha}}\to{\mathbb{R}}^{n}$ for each open set, ${\varphi}_{\alpha}$ is referred to as coordinate mapping and ${\mathbb{R}}^{n}$ is referred to as parameter domain.On the intersection ${\mathrm{U}_{\alpha}}\cap{\mathrm{U}_{\beta}}$, each point can have multiple local coordinates, and there is transformation ${\varphi}_{\alpha\beta}={\varphi}_{\beta}\circ{\varphi}_{\alpha}^{\mathrm{-1}}$ between local coordinates.The transformation ${\varphi}_{\alpha}:{\mathrm{U}_{\alpha}}\to{\mathbb{R}}^{n}$ from manifold to coordinate domain is called parameterization, its inverse transformation, and the transformation ${\varphi}_{\alpha}^{\mathrm{-1}}:{\varphi}_{\alpha}(\mathrm{U}_{\alpha})\to{\mathrm{U}_{\alpha}}$ from local coordinates to manifold is called the local parameter representation of manifold.If the manifold $\mathrm{S}$ is embedded in the Euclidean space ${\mathbb{R}}^{d}$, the Euclidean space ${\mathbb{R}}^{d}$ is called the background space.For example, the unit sphere embedded in three-dimensional Euclidean space is the simplest two-dimensional manifold.
The transformation from three-dimensional rectangular coordinate system to spherical coordinate system is called parameterized mapping, while the transformation from spherical coordinate system to three-dimensional rectangular coordinate system is called local parameter representation.

In deep learning, there are corresponding terms for these basic concepts of manifolds.
The parameter domain ${\mathbb{R}}^{n}$ is called the hidden space or the feature space;
Parameterized maps ${\varphi}_{\alpha}:{\mathrm{U}_{\alpha}}\to{\mathbb{R}}^{n}$ are called encoding maps;
The local parameter representation ${\varphi}_{\alpha}^{\mathrm{-1}}:{\varphi}_{\alpha}(\mathrm{U}_{\alpha})\to{\mathrm{U}_{\alpha}}$ of a manifold is referred to as a decoding map.
One of the main purposes and functions of deep learning is to learn encoding mapping and decoding mapping.
\begin{figure}[ht]
	\centering
	\includegraphics[scale=3,width=0.5\textwidth]{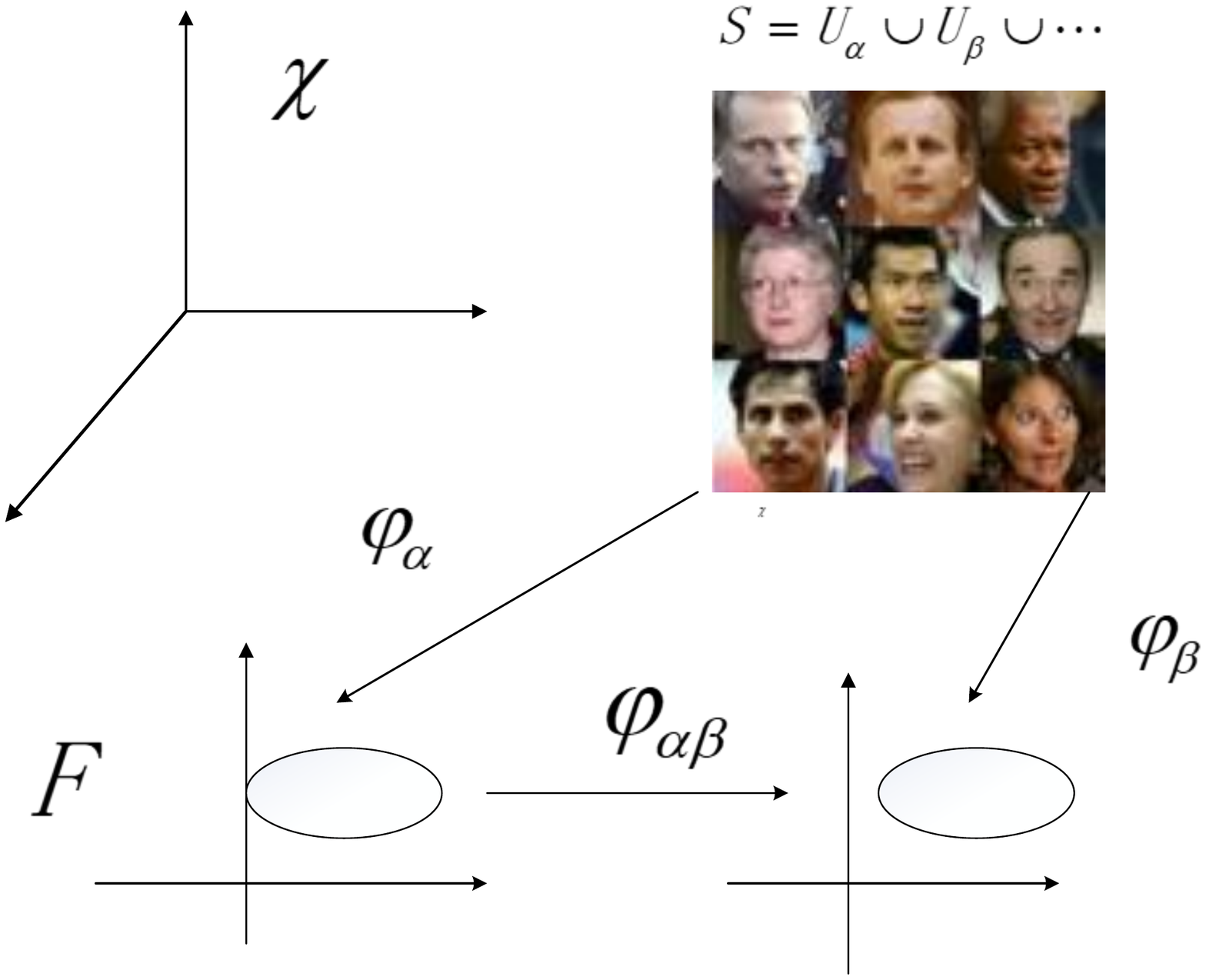}
	\caption{Manifold structure}
\end{figure}

\subsection{Encoder and decoder}
Autoencoder is a very basic deep learning model for learning manifold structures.
As shown in Figure 2, the automatic encoder is a feedforward network, with equal input and output dimensions, and both input and output are background spaces $\mathcal{X}$.
There is a bottleneck layer in the middle, and the output space of the bottleneck layer is the feature space $\mathcal{F}$.
The network is symmetric about the bottleneck layer, and the left network is used to represent the coding mapping, denoted as ${\varphi}_{\theta}:{\mathcal{X}}\to{\mathcal{F}}$;
The right network is used to represent decoding mapping, denoted as ${\psi}_{\theta}:{\mathcal{F}}\to{\mathcal{X}}$.
The loss function is equal to the norm $\mathrm{L}^{2}$ of the input and output images.
We take dense samples on manifolds, and get training sample set $\mathrm{X}=\{x_1,x_2,x_3,\dots,x_k\}$ to training network,
\begin{equation}
\min\limits_{\theta} \sum\limits_{i=1}^k\parallel{\mathrm{x_i}-{\psi}_{\theta}\circ{\varphi}_{\theta}(x_i)}\parallel^{2}
\end{equation}
We get encoding mapping and decoding mapping, decoding mapping is a parameter representation of manifold.
We use the reconstructed manifold $\mathrm{\widetilde{S}}:={\theta}\circ{\varphi}_{\theta}(\mathrm{S})$ to approximate data manifold $\mathrm{S}$.
\begin{figure}[ht]
	\centering
	\includegraphics[scale=3,width=0.5\textwidth]{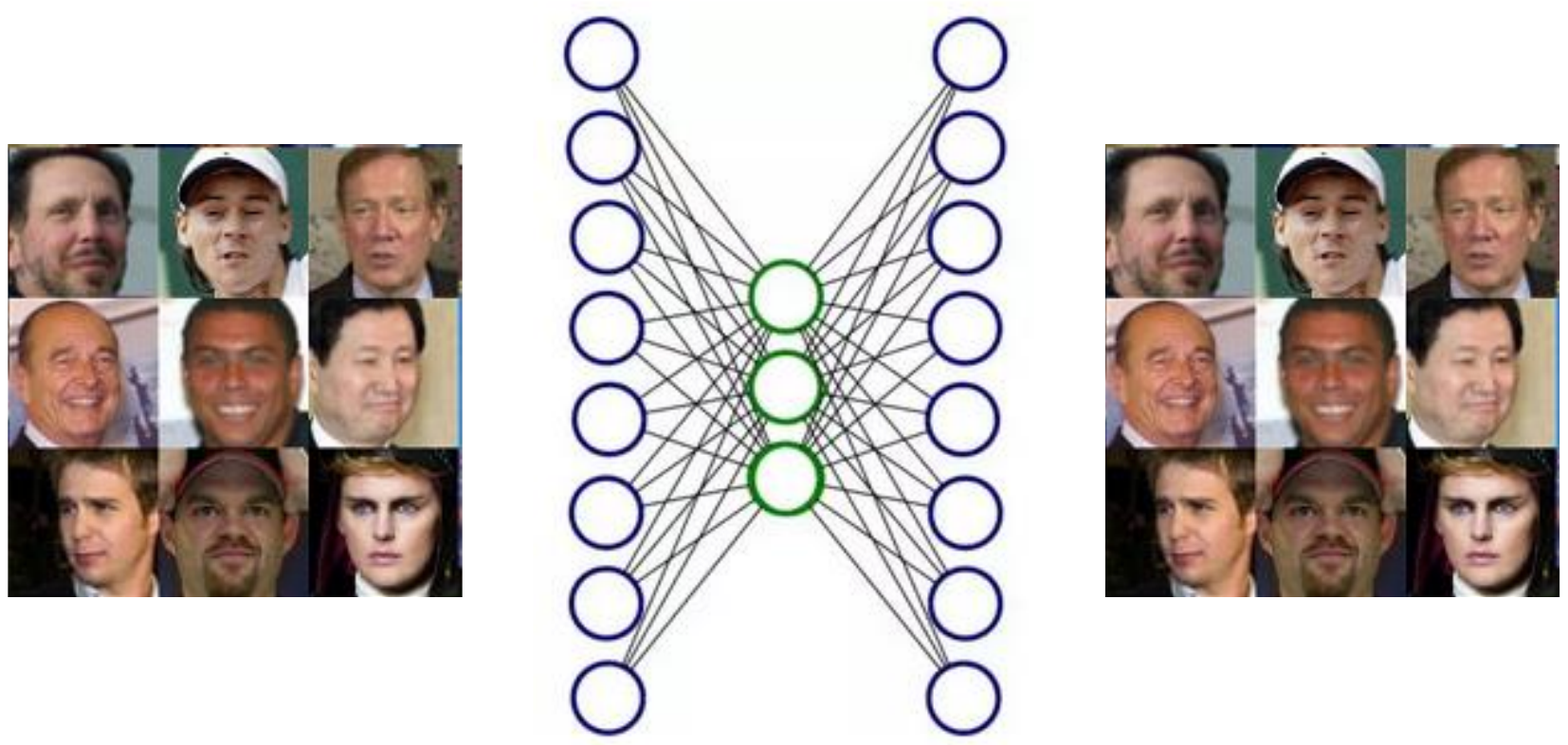}
	\caption{Encoder and decoder}
\end{figure}

\section{Model design and training of CNN}
In order to analyze the mechanism of convolution neural network, a neural network with shallow layer is designed in this section.The training of deep learning convolutional neural network is also a process of learning manifold structure. The process of forward propagation is essentially an encoding mapping, while its inverse process (back propagation or deconvolution visualization) is a decoding mapping.We explain this learning mechanism with the following experimental method.

\subsection{Model design}
The model adopts a typical feedforward neural network structure. As shown in Figure 1, the size of the data layer input image is 96*96. There are three convolution layers in the network. The number of convolution cores is 64, 128, 256 in turn. The design of gradually increasing convolution kernels is a common way. Its meaning is that with the increase of the number of layers, the features become more abstract and the high-level features formed by the combination of low-level features should be more and more. The size of each convolution core is 5*5. Each convolution layer is followed by a ReLU activation layer and a max pooling layer with size 2*2.Then connected to the full connection layer with 1024 hidden neurons, and finally connected to softmax layer as classification.

\begin{figure}[ht]
\centering
\includegraphics[scale=1,width=0.5\textwidth]{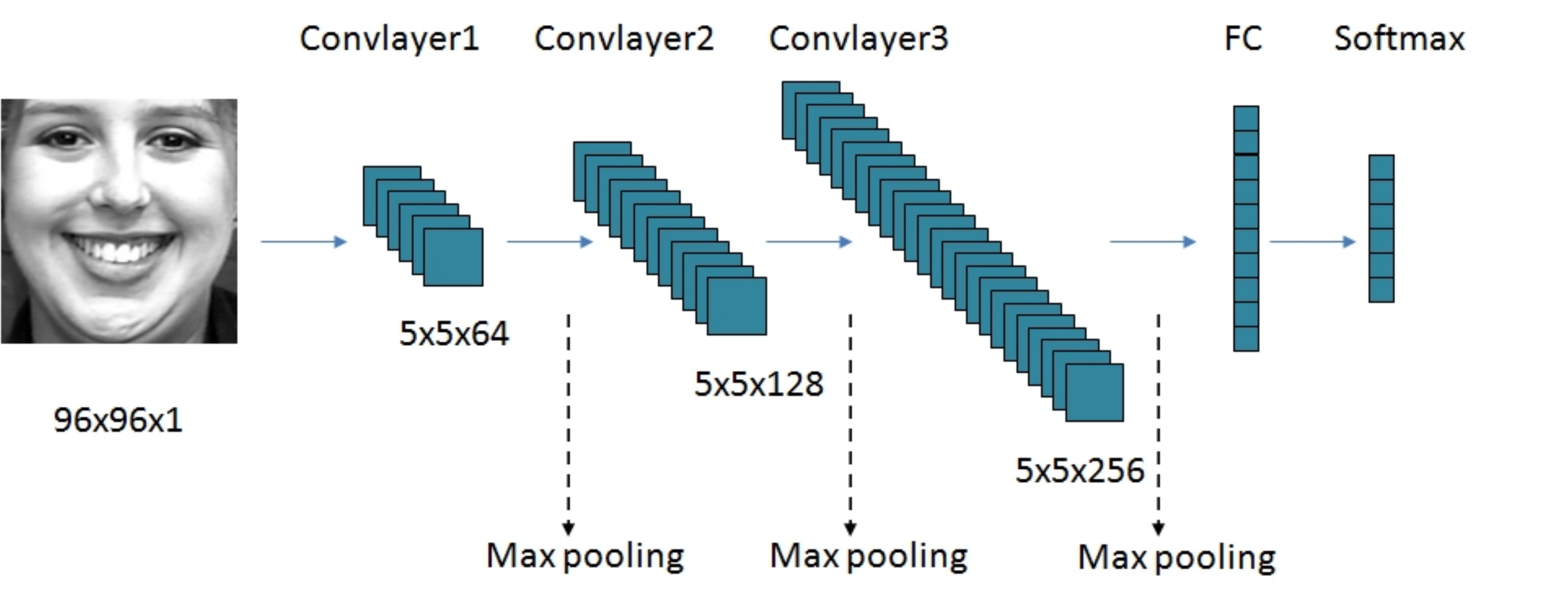}
\caption{Model network structure}
\end{figure}

\subsection{Model training}

The data set used in this experiment is: the extended Cohn-Kanade database (CK +)\cite{lucey2010extended}. The CK + dataset contains 327 face sequences whose tags contains the following seven expressions: anger, contempt, disgust, fear, joy, sadness, and surprise. The tag of each face picture has both the expression tag and the action unit tag which constitutes the expression.

This is a new network. So there are no pre-trained model parameters. The hyperparameters of the model are as follows: batchsize is set to 64, momentum is set to 0.9, weight decay parameter is 0.0001, learning rate is constant 0.001, weight is initialized by Gaussian distribution with mean value 0 and variance 1. In order to avoid over-fitting of the model, the method of data enhancement and dropout is adopted in this paper. The parameter for the full connection layer is dropout with a probability value of 0.5.

To increase the fairness of comparison, this section uses the same data sets and tags as some previous work. The first frame of each picture sequence is treated as a neutral expression and the other pictures are treated as data sets of the training model. In the enhanced data set, 200 images are selected as the test set. The remaining images are used as the training set. The expression picture here is not the original picture, but a picture of the face area. According to the coordinates of the feature points of each image given by the ck + data set, the area surrounded by the upper, lower, left and right four largest feature points is selected as the face image.

For the sake of fairness, this section trains the network from scratch. Data preprocessing and enhancement are as follows:
\begin{enumerate}
\item Rotate the picture randomly between angles (-15, + 15).
\item Flip the picture at random.
\item Normalize the image to the size of 99*99.
\item Cut the picture randomly to 96*96.
\item Standardize the picture, that is, subtract the average value of the picture pixel from each pixel of the picture, and divide by the standard deviation of the picture pixel.
\end{enumerate}

Training the model until the model converges.

\subsection{Brief  of  traditional expression recognition method}
One of the traditional expression recognition methods is based on action unit (AU). The action unit is derived from the Facial Action Coding System, which is for dividing the movement regions of facial muscles. It was developed by Paul Ekman and Wallace in 1978 \cite{ekmanfacial}. The movements of each facial muscle can be coded to represent changes in the appearance of the face. The coding can be used as a standard for systematic expression classification.

The unit of FACS is the Action Unit (AU). The action unit is a recognizable facial change caused by the simultaneous action of several muscles, see Table 1. Almost all facial expressions can be broken down into combinations of different facial units, see Figure 2.

\begin{figure}[ht]
\centering
\includegraphics[scale=1,width=0.5\textwidth]{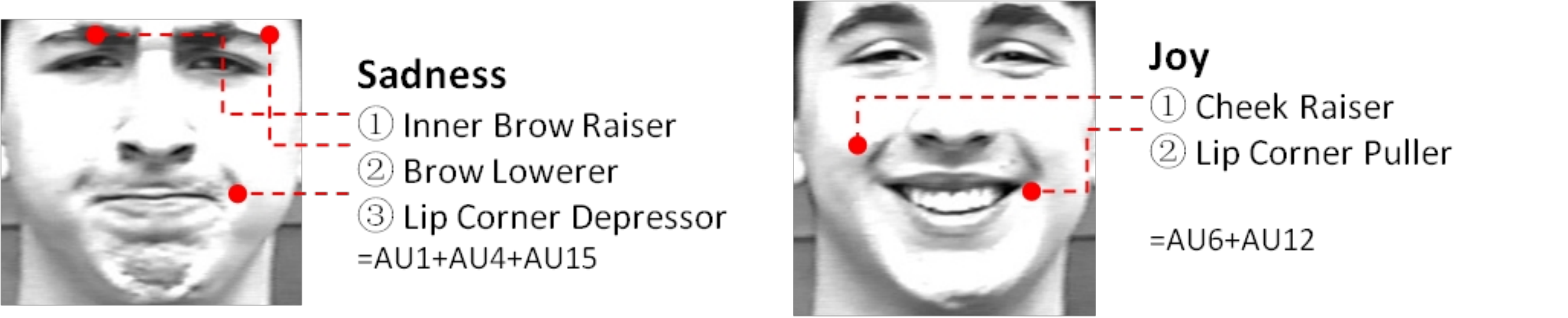}
\caption{Expression coding}
\end{figure}

In traditional methods, it is common to design a detector for each facial action unit. The expressions are classified according to the detection results of the monitor, see Figure 3.

\begin{figure}[ht]
\centering
\includegraphics[scale=1,width=0.5\textwidth]{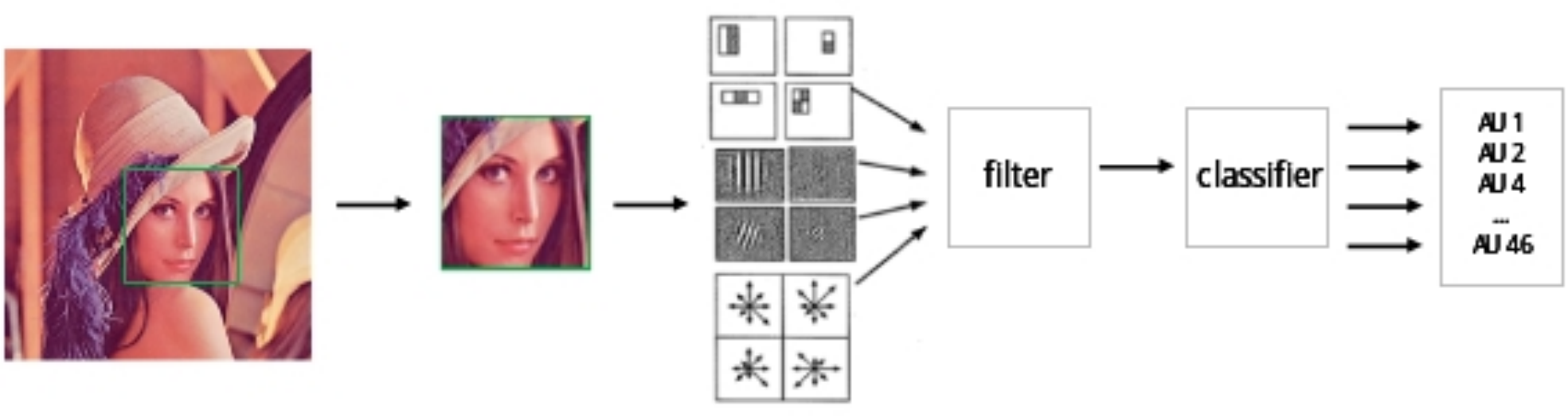}
\caption{Traditional method to detect facial action unit.}
\end{figure}

This method is theoretically feasible. However, in practice, it is very difficult to design the detector of each action unit manually and ensure its accuracy and efficiency. The reason that neural network can classify exactly is that it can learn the feature automatically from the sample, which saves the process of extracting the feature manually. As to the principle of neural network classification model, this paper proposes the following hypothesis: the trained convolution neural network learns the detector similar to facial action unit.

\section{Analyzing the classification mechanism of the network}
\subsection{Deconvolution visualization of model}
In section 2, we obtain a better model for the expression classification of the training set. The model can effectively classify the expressions in the training set. Taking the model as the research object, the classification mechanism of the convolution model is discussed in this paper. In this section, we use deconvolution visualization\cite{zeiler2014visualizing} technology to map the largest excitation response in the feature map of different layers to the original picture pixel space, and explore the regions which have the most important impact on the model classification. As a diagnostic tool, deconvolution visualization enables us to understand the role played by each layer and the potential problems of the diagnostic model.

It is a common practice to obtain a visual understanding of network functions through visual features\cite{donahue2014decaf}, but it is usually limited to the first layer of the network directly mapped to the image. For the response of a neuron in a high-level feature, to map back to its corresponding pixel space, it needs to be realized by deconvolution network\cite{erhan2009visualizing}. For example, in order to visualize the n-layer neuron activation value a, it is first necessary for the image to be propagated forward through the network. Next, the n-layer neurons other than the activation value a are set to 0, and then the image of the neurons activated in the original pixel space is visualized through deconvolution.

Let the input image be x , the convolution core C, and the output be y. The convolution operation can be represented as:

\begin{equation}
\mathrm{y=Cx}
\end{equation}

In reverse propagation, you get $ \frac{\partial{Loss}}{\partial{y}}$ from a deeper network.

\begin{gather}
\frac{\partial\mathrm{Loss}}{\partial{\mathrm{x}}_{\mathrm{j}}}\mathrm{=}\sum_{\mathrm{i}}{\frac{\partial\mathrm{Loss}}{\partial{\mathrm{y}}_{\mathrm{i}}}}\frac{\partial{\mathrm{y}}_{\mathrm{i}}}{\partial{\mathrm{x}}_{\mathrm{j}}}\mathrm{=}\sum_{\mathrm{i}}{\frac{\partial\mathrm{Loss}}{\partial{\mathrm{y}}_{\mathrm{i}}}}{\mathrm{C}}_{\mathrm{i,j}}\mathrm{=}{\mathrm{C}}^{\mathrm{T}}_{\mathrm{*}\mathrm{,j}}\frac{\partial\mathrm{Loss}}{\partial\mathrm{y}}
\end{gather}

\begin{equation}
\frac{\partial\mathrm{Loss}}{\partial\mathrm{x}}\mathrm{=}\left[ \begin{array}{c}
\frac{\partial\mathrm{Loss}}{\partial{\mathrm{x}}_{\mathrm{1}}} \\ 	
\frac{\partial\mathrm{Loss}}{{\partial\mathrm{x}}_{\mathrm{2}}} \\
... \\
\frac{\partial\mathrm{Loss}}{\partial{\mathrm{x}}_{\mathrm{n}}} \end{array}
\right]\mathrm{=}\left[ \begin{array}{c}
{\mathrm{C}}^{\mathrm{T}}_{\mathrm{*}\mathrm{,1}}\frac{\partial\mathrm{Loss}}{\partial\mathrm{y}} \\
{\mathrm{C}}^{\mathrm{T}}_{\mathrm{*}\mathrm{,2}}\frac{\partial\mathrm{Loss}}{\partial\mathrm{y}} \\
... \\
{\mathrm{C}}^{\mathrm{T}}_{\mathrm{*}\mathrm{,n}}\frac{\partial\mathrm{Loss}}{\partial\mathrm{y}} \end{array}
\right]\mathrm{=}\left[ \begin{array}{c}
{\mathrm{C}}^{\mathrm{T}}_{\mathrm{*}\mathrm{,1}} \\
{\mathrm{C}}^{\mathrm{T}}_{\mathrm{*}\mathrm{,2}} \\
... \\
{\mathrm{C}}^{\mathrm{T}}_{\mathrm{*}\mathrm{,2}} \end{array}
\right]\frac{\partial\mathrm{Loss}}{\partial\mathrm{y}}\mathrm{=}{\mathrm{C}}^{\mathrm{T}}\frac{\partial\mathrm{Loss}}{\partial\mathrm{y}}
\end{equation}

In this section, we use deconvolution visualization method to visualize the active values on the corresponding part of the feature graph of the third layer of convolution layer in the network.
The deconvolution of the activation value of the feature graph is shown in Figure 4. The active value i is an active value in the feature graph j generated by the third convolution layer. By deconvolution of the active value i, the receptive field (block a) mapped on the original graph and the corresponding deconvolution response (block b) can be obtained. From the deconvolution process of formula 1-3, it is known that the graph represented by block b can increase the value of the active value i. Therefore, it can be considered that the network structure associated with the feature graph j in which the active value i is located constitutes a detector of graph block b.

\begin{figure}[ht]
\centering
\includegraphics[scale=1,width=0.5\textwidth]{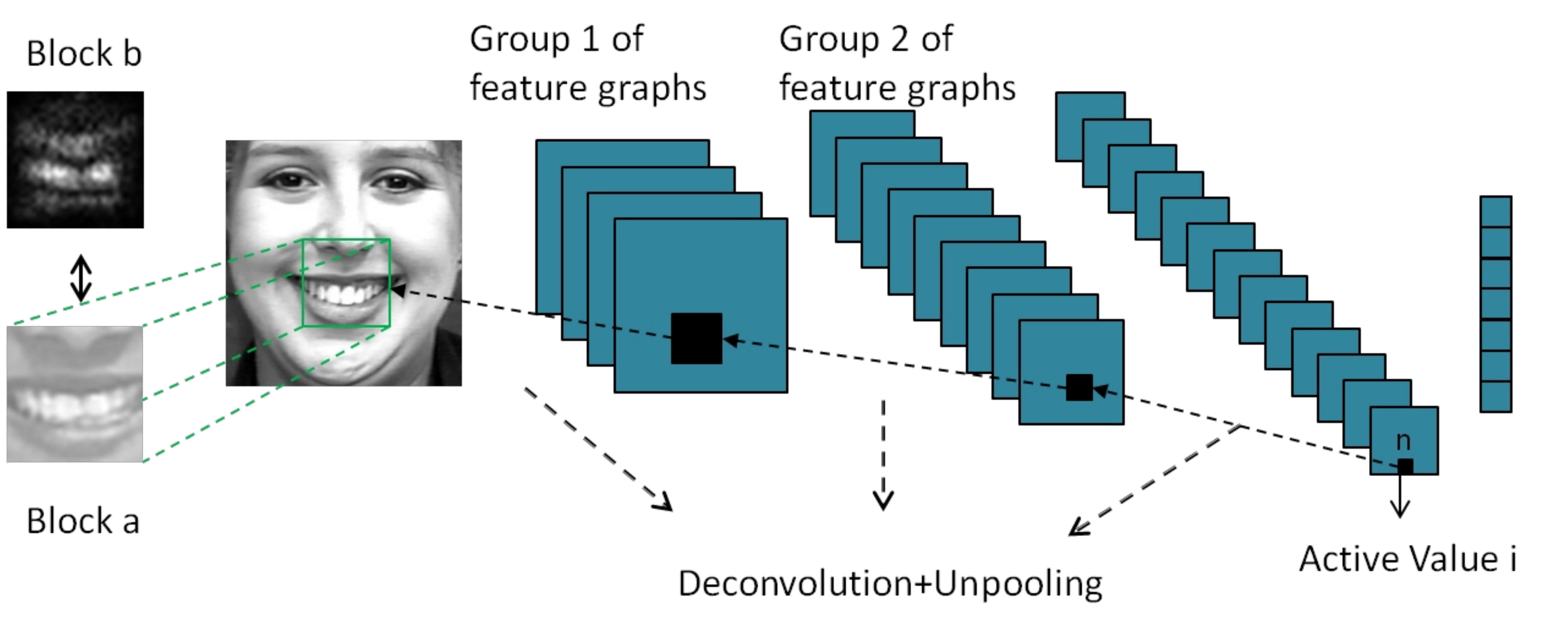}
\caption{Deconvolution visualization of active values.}
\end{figure}

\subsection{Facial feature units}
The process of this experiment is as follows:
\begin{enumerate}
\item Enter all the training pictures into the network in turn.
\item For each picture, record the maximum activation value and the corresponding position of each picture in 256 characteristic graphs in the third characteristic graphs.
\item For each feature image, take the picture corresponding to the first nine maximum activation values.
\item Using deconvolution network, the response of original image space corresponding to each activation value is calculated.
\item At the same time, the activation value is calculated in the corresponding area of the original image.
\end{enumerate}
Finally, the responses of the receptive field and deconvolution corresponding to the first nine maximum values of each feature graph in the third layer are obtained. Some deconvolution visualizations of the feature graphs are selected, and the effect is shown in Table 2.

\newcommand{\tabincell}[2]{\begin{tabular}{@{}#1@{}}#2\end{tabular}}
\begin{table*}[ht]
\centering
\caption{Comparison of deconvolution response graphs with facial feature units}
\begin{tabular}{cccc}
	\toprule
	\bf{AU}&Picture AU&Type&\tabincell{c}{The response of deconvolution of the maximum\\activation value in the feature graph}\\
	\midrule
	1 & \includegraphics[height=0.25in]{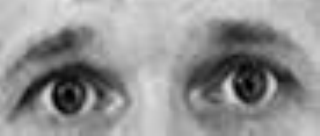} & Original& \includegraphics[height=0.25in]{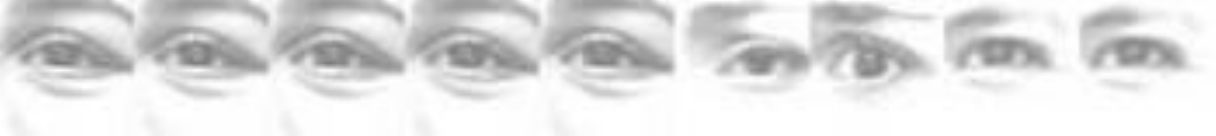}\\
	&   & Deconvolution& \includegraphics[height=0.25in]{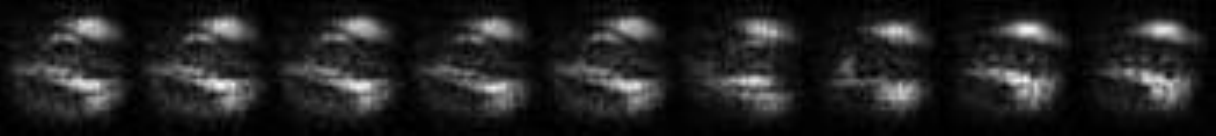}\\
	
	16 & \includegraphics[height=0.25in]{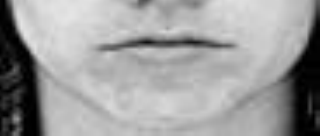} & Original& \includegraphics[height=0.25in]{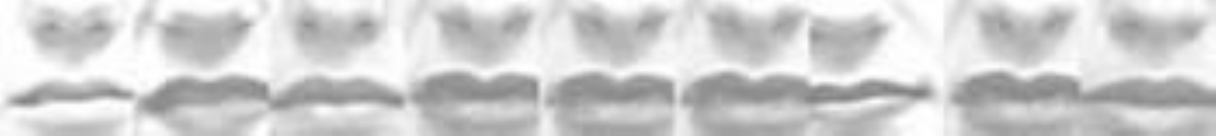}\\
	&   & Deconvolution& \includegraphics[height=0.25in]{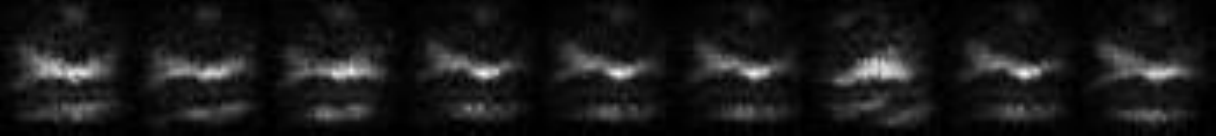}\\
	
	6 & \includegraphics[height=0.25in]{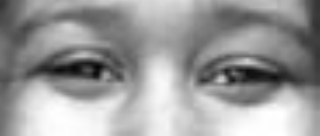} & Original& \includegraphics[height=0.25in]{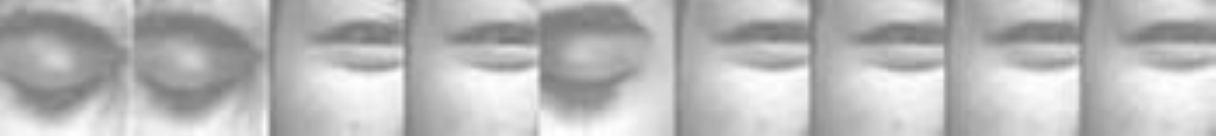}\\
	&   & Deconvolution& \includegraphics[height=0.25in]{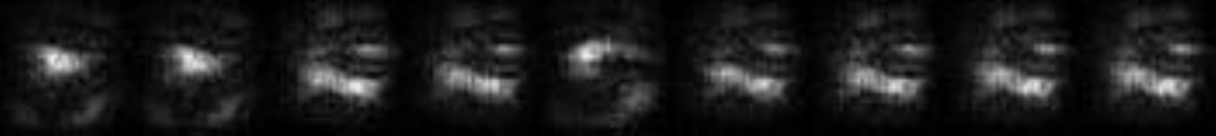}\\
	
	26 & \includegraphics[height=0.25in]{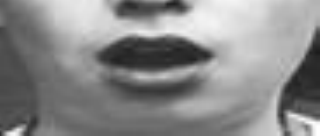} & Original& \includegraphics[height=0.25in]{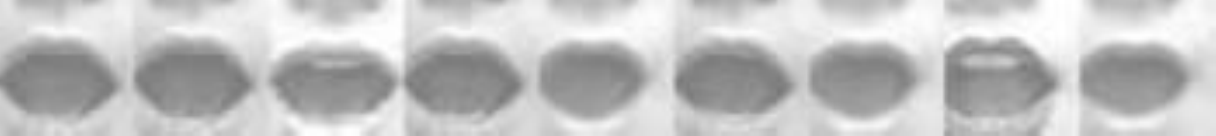}\\
	&   & Deconvolution& \includegraphics[height=0.25in]{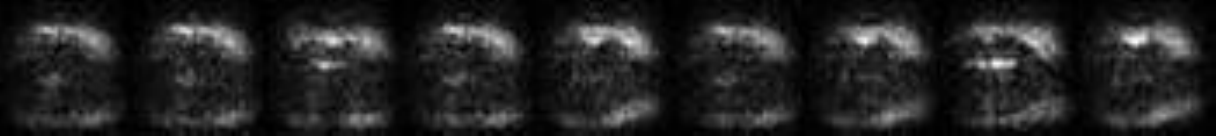}\\
	
	2 & \includegraphics[height=0.25in]{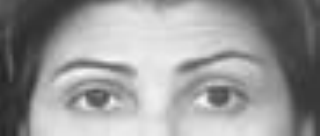} & Original& \includegraphics[height=0.25in]{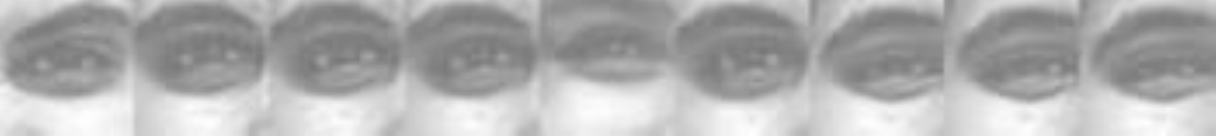}\\
	&   & Deconvolution& \includegraphics[height=0.25in]{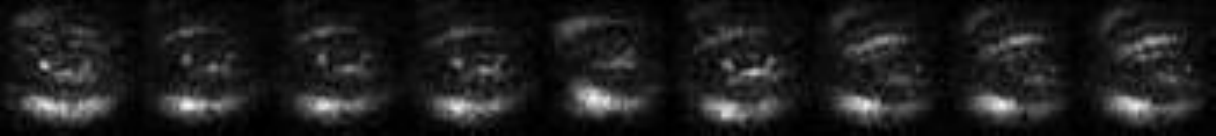}\\
	
	27 & \includegraphics[height=0.25in]{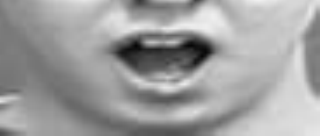} & Original& \includegraphics[height=0.25in]{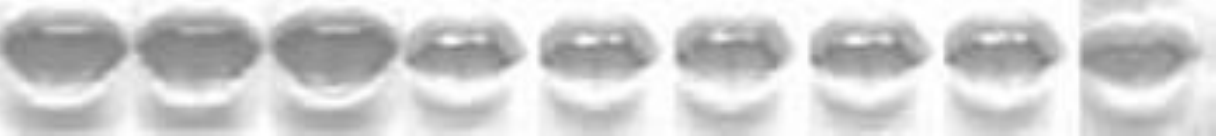}\\
	&   & Deconvolution& \includegraphics[height=0.25in]{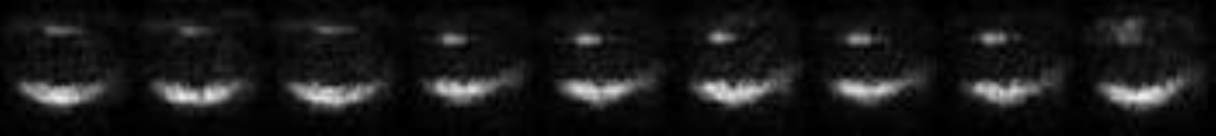}\\
	
	\bottomrule
\end{tabular}
\end{table*}

As shown in Table 2, these are the response of the original image space corresponding to the activation value of the odd behavior characteristic graph after deconvolution and the image receptive field corresponding to the even behavior. It can be clearly seen that: 1.The image area corresponding to the maximum response value of a characteristic graph of the trained network is basically the same image type. And its deconvolution corresponding to the activation pattern in the original image pixel space is also very similar to the image type. 2. The image obtained by partial deconvolution is very close to the facial feature unit.

Therefore, the following preliminary conclusions can be drawn: through training the network model, the model gradually formed some specific shape of the detector. Some of the shapes detected are very similar to those of facial motion units.

\subsection{Mouth shape in deconvolution feature graph}
The number of feature graphs outputted from the third layer is 256, which is equivalent to 256 feature detectors. The number of these feature detectors is far more than the number of facial action units. So the division of facial regions is more detailed. Table 3 is the mouth shape selected from the deconvolution feature map. And Table 4 is the mouth shape defined by the facial action unit.

\begin{table}[ht]
\centering
\caption{mouth shape in deconvolution feature graph}
\begin{tabular}{ccc}
	\toprule
	\bf{id}&Type&\tabincell{c}{Deconvolution example}\\
	\midrule
	1  & Original& \includegraphics[height=0.25in]{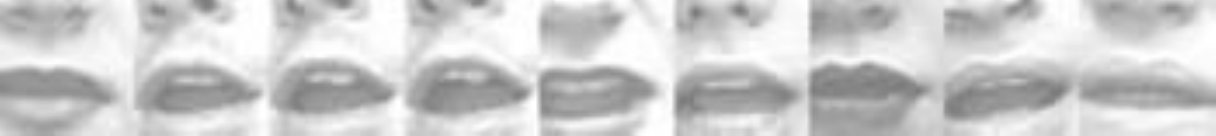}\\
	& Deconv& \includegraphics[height=0.25in]{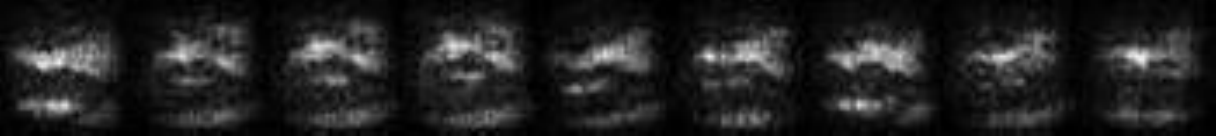}\\
	2  & Original& \includegraphics[height=0.25in]{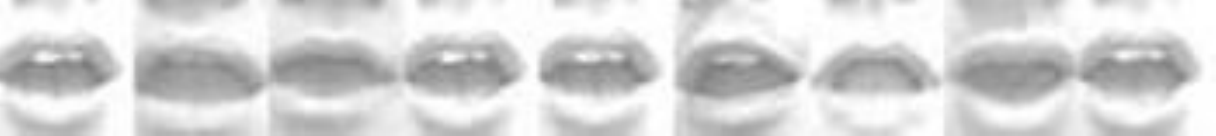}\\
	& Deconv& \includegraphics[height=0.25in]{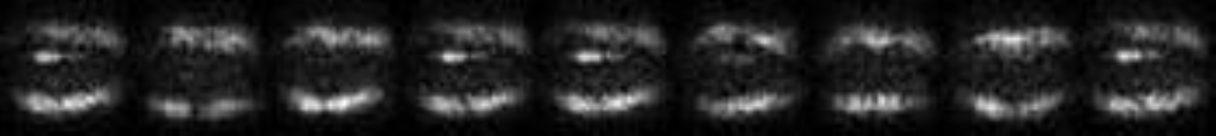}\\
	3  & Original& \includegraphics[height=0.25in]{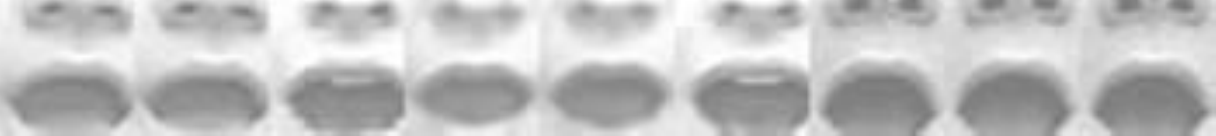}\\
	& Deconv& \includegraphics[height=0.25in]{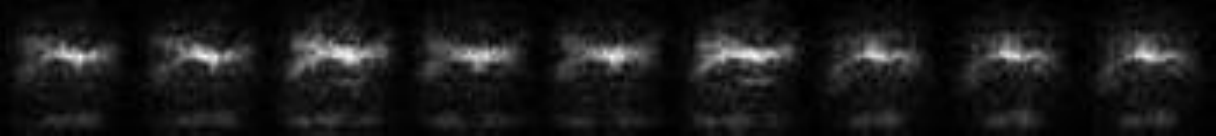}\\
	4  & Original& \includegraphics[height=0.25in]{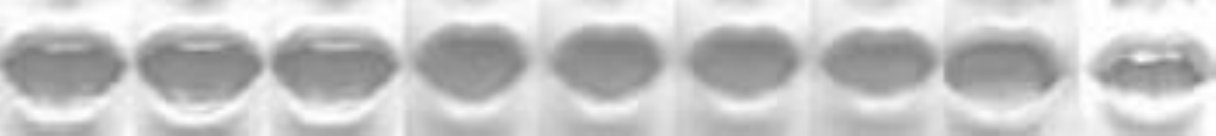}\\
	& Deconv& \includegraphics[height=0.25in]{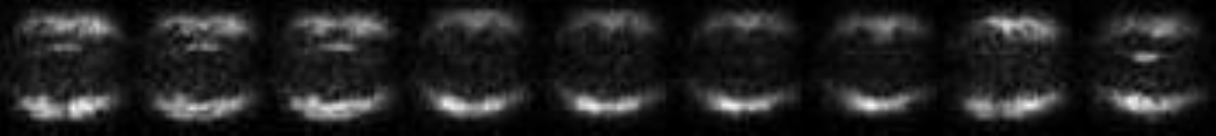}\\
	5  & Original& \includegraphics[height=0.25in]{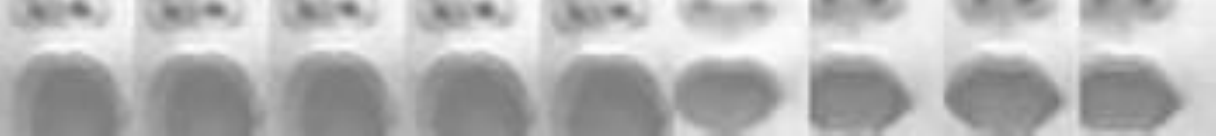}\\
	& Deconv& \includegraphics[height=0.25in]{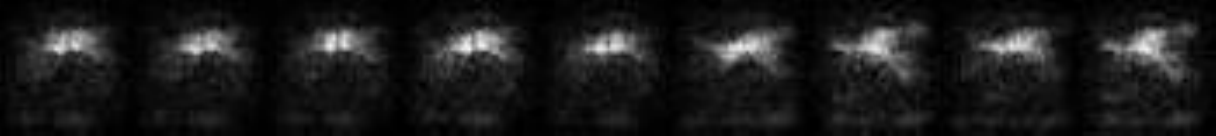}\\
	6  & Original& \includegraphics[height=0.25in]{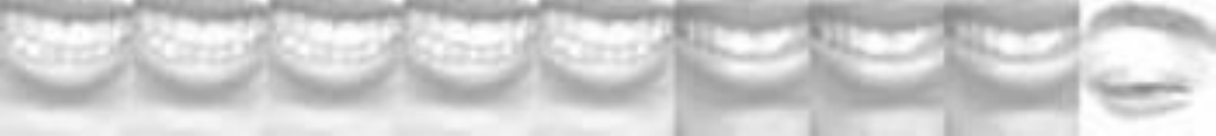}\\
	& Deconv& \includegraphics[height=0.25in]{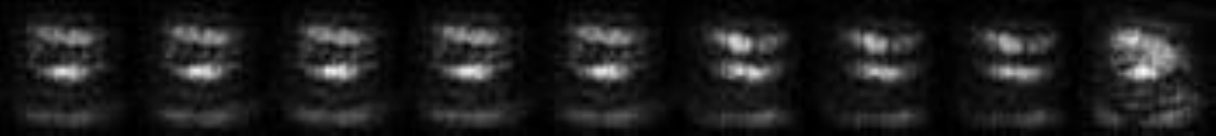}\\
	
	\bottomrule
\end{tabular}
\end{table}

\begin{table}[ht]
\centering
\caption{mouth shapes in facial action units}
\begin{tabular}{cc}
	\toprule
	\midrule
	Lips apart & \includegraphics[height=0.3in]{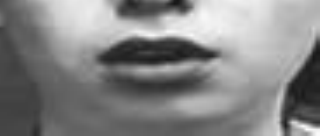}\\
	Jaw Drop & \includegraphics[height=0.3in]{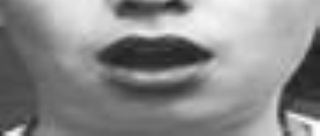}\\
	Mouth Stretch & \includegraphics[height=0.3in]{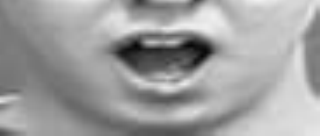}\\
	\bottomrule
\end{tabular}
\end{table}

It can be seen that the trained network can detect a variety of facial mouth deformation. Compared with the three mouth shapes defined in the facial action unit, the convolution network detects more types of mouth shapes. So the expression classification based on convolution network is more delicate than the facial expression classification based on facial motion unit.

Compared with the process of designing the detector by hand, the method of deep learning automatically completes the extraction of data features in the process of training the network and forms a detector for specific graphics. Data-driven algorithm instead of manual design to extract features makes the problem easier and faster processing, which is a great advantage of deep learning methods.

This section presents the internal mechanism of the network with deconvolution visualization. The similarity between the network-formed detector and the facial motion unit is also visualized. And followed we demonstrate the relationship between the detector formed by the network and the facial motion unit.

\section{Validation of Correlation between Deconvolution Feature Map and Facial Action Units}
The deconvolution feature graphs of the network are obtained in the preceding section. And we can observe that some of the patterns are very similar to the facial feature units. In this section, the correlation between deconvolution feature image and facial feature unit is verified. The one-to-one correspondence between partial deconvolution feature map and facial feature unit is proved. That is to say, for a certain facial action unit, the most closely associated feature graph is obtained.
\subsection{The formula of best simile between Feature Map and Unit}
This section proposes the following hypothesis: for a particular feature graph i in the network, let the feature graph represent the detector of the facial action unit j. S is a collection of pictures containing action unit j and $S^c$ is a collection of pictures without action unit j. Let S respond through the network in the characteristic graph a as
$F_{a}(S) $. $S^c$ respond through the network in the characteristic graph a as $F_{a}(S^c) $.And the distance between $F_{a}(S)$ and $F_{a}(S^c)$ is $D(F_{a}(S),F_{a}(S^c))$.There is:
\begin{equation}
\mathrm{D(F_{i}(S),F_{i}(S^c))=max(D(F_{a}(S)),F_{a}(S^c)) \quad a =1,...,256 }
\end{equation}
That is, the maximum distance between $F_{a}(S)$ and $F_{a}(S^c)$ should be $D(F_{i}(S),F_{i}(S^c))$. This hypothesis is reasonable.Because in general, for the detector i of the action unit j , if S contains an action unit j and the response $F_{i}(S)$ is larger.Otherwise the response $F_{i}(S^c)$ is smaller. Therefore, the distance $D(F_{i}(S),F_{i}(S^c))$ should be the maximum of the distance in all the characteristic graphs for the characteristic graphs i.

Therefore, for a given facial action unit j, the feature graph i that maximizes $D(F_{a}(S),F_{a}(S^c))$  is the feature graph that is most closely associated with the facial action unit j. That is, the network structure associated with the feature graph i constitutes a detector for the facial feature unit j.

The experimental process is as follows:

In addition to providing facial expression tags, the ck+ database also provides FAU tags for images. Through the FAU tag, the correlation between the spatial pattern of the original image (i.e. deconvolution feature map) and the action unit can be verified. The activation value distribution of the picture with action unit j is the most different from that of the picture without action unit j in the feature map i.

Let $F_{Li(x)}$ be the activation value of the first i feature graph of picture x in the L layer.Here we use the third layer convolution kernel, so L=3. S is a collection of pictures containing action unit j and $S^c$ is a collection of pictures without action unit j.

\begin{equation}
{\mathrm{R}}_{\mathrm{ij}}\left(\mathrm{x}\right)\mathrm{=P(}{\mathrm{F}}_{\mathrm{3i}}\mathrm{(x|S))}
\end{equation}
\begin{equation}
{\mathrm{Q}}_{\mathrm{ij}}\left(\mathrm{x}\right)\mathrm{=P(}{\mathrm{F}}_{\mathrm{3i}}\mathrm{(x|}{\mathrm{S}}^{\mathrm{c}}\mathrm{))}
\end{equation}

$Q_{ij}(x)$ and $R_{ij}(x)$ are the probability distributions on the characteristic graph i of the action unit with j and the action unit without j respectively. The common method for measuring the distance between two distributions is to calculate the KL divergence between the two distributions.

KL distance \cite{tibshirani1987local} (relative entropy): A measure of the difference between the two probability distributions of the same event space. The formula is as follows:
\begin{equation}
\mathrm{D(R|}\left|\mathrm{Q}\right)\mathrm{=}\sum_{\mathrm{x}\mathrm{\in }\mathrm{X}}{\mathrm{R(x)log}\mathrm{}\mathrm{(}\frac{\mathrm{R(x)}}{\mathrm{Q(x)}}\mathrm{)}}
\end{equation}

When R and Q are the same distribution, the KL distance is 0. And the greater the difference of R and Q distribution, the greater the KL distance. In this section, the distance between $Q_{ij}(x)$ and $R_{ij}(x)$ does not strictly conform to the definition of KL distance: 1.The probability distributions represented by $Q_{ij}(x)$ and $R_{ij}(x)$ cannot be obtained. 2. $Q_{ij}(x)$ and $R_{ij}(x)$ are input from different data sets.

To solve the above problems, the following distance function is designed in this section :

Step 1:Let the activation value of the feature graph j be $R_{j}(x)$ and $Q_{j}(x^c)$,where x belongs to the first n pictures in which the activation value response of the image set S is greatest in the feature graph j, and $x^c$ belongs to the first n pictures in which the activation value response of the image set $S^c$ is greatest in the feature graph j.There are:
\begin{equation}
{\mathrm{R}}_{\mathrm{j}}\left({\mathrm{x}}_{\mathrm{1}}\right)\mathrm{>}{\mathrm{R}}_{\mathrm{j}}\left({\mathrm{x}}_{\mathrm{2}}\right)\mathrm{>}...{\mathrm{>}R}_{\mathrm{j}}\left({\mathrm{x}}_{\mathrm{n}}\right)
\end{equation}
\begin{equation}
{\mathrm{Q}}_{\mathrm{j}}\left({{\mathrm{x}}^{\mathrm{c}}}_{\mathrm{1}}\right)\mathrm{>}{\mathrm{Q}}_{\mathrm{j}}\left({{\mathrm{x}}^{\mathrm{c}}}_{\mathrm{2}}\right)\mathrm{>}...{\mathrm{>}Q}_{\mathrm{j}}\left({{\mathrm{x}}^{\mathrm{c}}}_{\mathrm{n}}\right)
\end{equation}
These activation values $P_{j}(x)$ and $Q_{j}(x^c)$ as follows:
\begin{equation}
\mathrm{D(R|}\left|\mathrm{Q}\right)\mathrm{=}\sum^{\mathrm{n}}_{\mathrm{k=1}}{{\mathrm{R}}_{\mathrm{j}}\left({\mathrm{x}}_{\mathrm{k}}\right){\mathrm{log} \left(\frac{{\mathrm{R}}_{\mathrm{j}}\left({\mathrm{x}}_{\mathrm{k}}\right)}{{\mathrm{Q}}_{\mathrm{j}}\left({{\mathrm{x}}^{\mathrm{c}}}_{\mathrm{k}}\right)}\right)\ }}
\end{equation}

Its property is similar to that of KL distance. The closer the response of two data sets in the feature graph j (i.e.$R_{j}(x)$ and $Q_{j}(x^c)$) are,the smaller the distance is, and vice versa. Because the distance is asymmetrical, i.e:
\begin{equation}
\mathrm{D(R|}\left|\mathrm{Q}\right)\mathrm{¡Ù}\mathrm{D(Q|}\left|\mathrm{R}\right)
\end{equation}
So the final distance function is as follows:
\begin{equation}
{\mathrm{D}}_{\mathrm{j}}\mathrm{(R|}\left|\mathrm{Q}\right)\mathrm{=D(R|}\left|\mathrm{Q}\right)\mathrm{+D(Q|}\left|\mathrm{R}\right)
\end{equation}
\begin{equation}
{\mathrm{D}}_{\mathrm{j}}\mathrm{(R|}\left|\mathrm{Q}\right)\mathrm{=}\sum^{\mathrm{n}}_{\mathrm{k=1}}{\mathrm{R}\left({\mathrm{x}}_{\mathrm{k}}\right){\mathrm{log} \left(\frac{{\mathrm{R}}_{\mathrm{j}}\left({\mathrm{x}}_{\mathrm{k}}\right)}{{\mathrm{Q}}_{\mathrm{j}}\left({{\mathrm{x}}^{\mathrm{c}}}_{\mathrm{k}}\right)}\right)\ }}\mathrm{+}\sum^{\mathrm{n}}_{\mathrm{k=1}}{{\mathrm{Q}}_{\mathrm{j}}\left({{\mathrm{x}}^{\mathrm{c}}}_{\mathrm{k}}\right){\mathrm{log} \left(\frac{{\mathrm{Q}}_{\mathrm{j}}\left({{\mathrm{x}}^{\mathrm{c}}}_{\mathrm{k}}\right)}{{\mathrm{R}}_{\mathrm{j}}\left({\mathrm{x}}_{\mathrm{k}}\right)}\right)\ }}
\end{equation}
The distance of the facial action unit i on the feature graph j is
In this experiment, we select n=9. That is, the first nine images with the greatest response value on the feature graph j are selected for the data set containing the facial action unit i and the data set without the facial action unit i respectively. And the distance is calculated as:
\begin{equation}
{\mathrm{D=D}}_{\mathrm{ij}}\mathrm{(}{\mathrm{Q}}_{\mathrm{ij}}\mathrm{||}{\mathrm{R}}_{\mathrm{ij}}\mathrm{)}
\end{equation}
The distance is calculated for all 256 feature maps and the corresponding feature map with the maximum distance is selected:
\begin{equation}
{\mathrm{max}\mathrm{}\mathrm{(D}}_{\mathrm{ij}}\mathrm{(}{\mathrm{Q}}_{\mathrm{ij}}\mathrm{||}{\mathrm{R}}_{\mathrm{ij}}\mathrm{)),\ \ \ j=1,}...\mathrm{,256}
\end{equation}
When j=1(i.e. choosing FAU1), the distance $D_{ij}$ corresponding to each feature graph is demonstrated as figure 5.
\begin{figure}[ht]
\centering
\includegraphics[scale=1,width=0.5\textwidth]{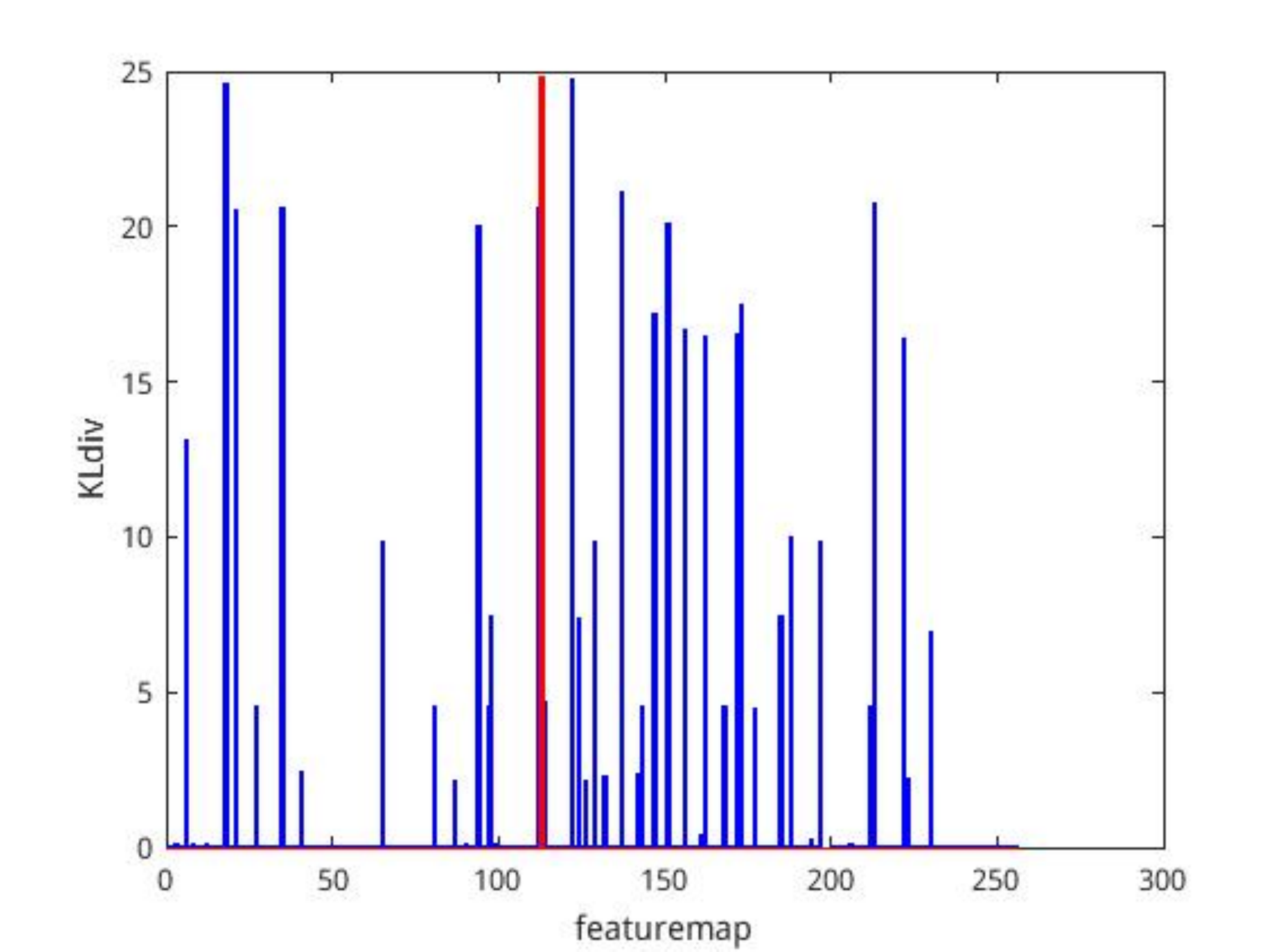}
\caption{The distance $D_{i}$ of FAU1 mapped on the third feature graphs.}
\end{figure}

\subsection{Results}
Figure 5 shows that the distance of FAU1 is the largest when i=113. And the deconvolution feature map and the receptive field region of the original map are shown in Table 4.

It is evident from table 4 that the face action unit 1 is highly consistent with the deconvolution image of feature map 113. At this point, we obtain the mapping relationship between the facial action unit 1 and the feature graph 113 and draw the following conclusion: the network structure associated with the feature graph 113 is the detector of the facial action unit 1. In this section, by calculating the distance $D_{j}$ of activation value distribution of facial action unit j in different feature graphs, the mapping between facial action unit and feature graphs is found successfully. Figure 6 is a distance image on a third layer feature map of a portion of the AU unit.
\begin{figure}[h]
\centering
\includegraphics[scale=1,width=0.5\textwidth]{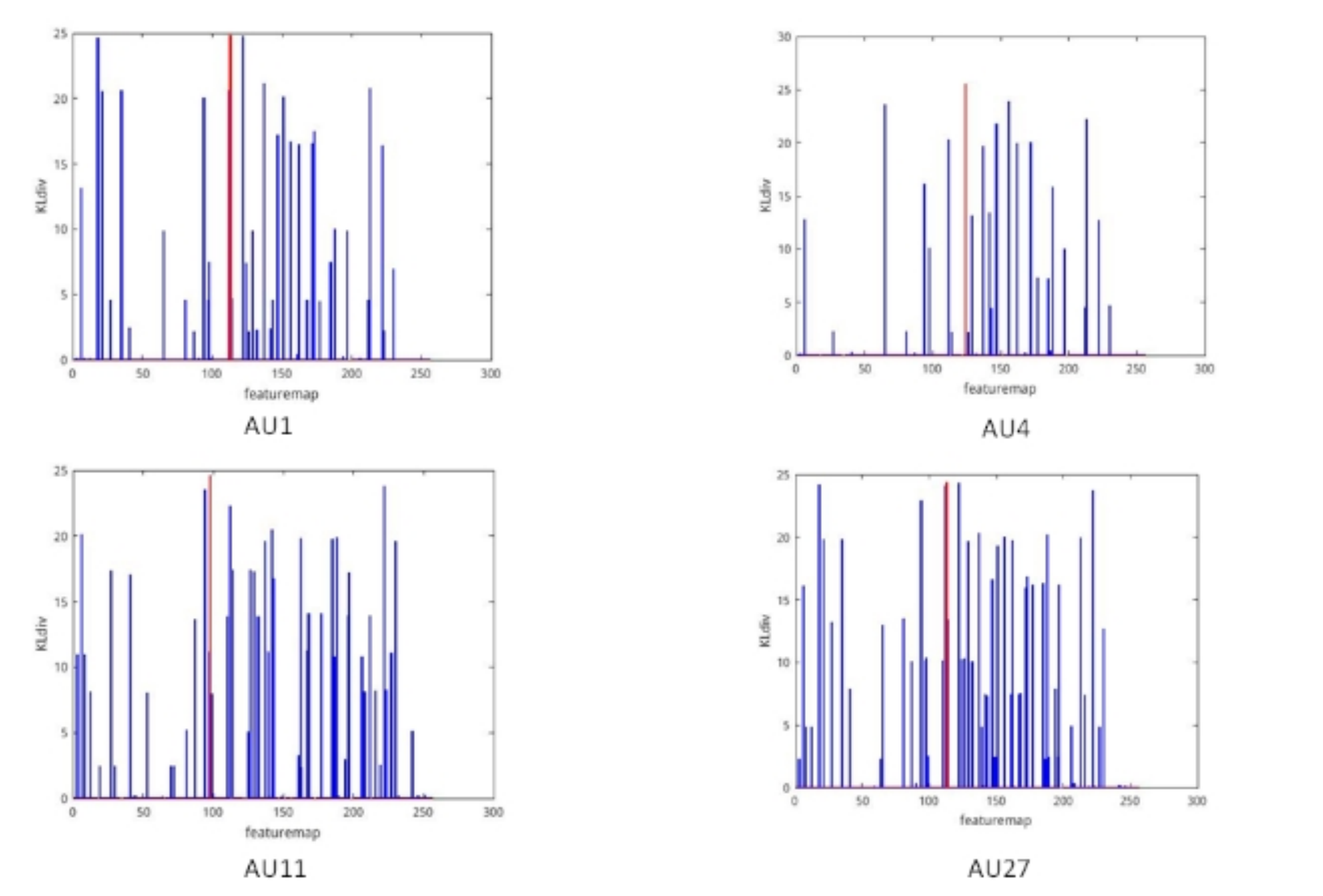}
\caption{The distance images corresponding to AU1, AU4, AU11, AU27.}
\end{figure}

Table 5 is a deconvolution feature map of the feature map with the largest distance corresponding to the facial feature unit of figure 6.
\begin{table}[h]
\centering
\caption{Comparison between deconvolution feature graph of 113 and FAU1}
\begin{tabular}{p{0.5cm}p{0.5cm}ccc}
	\toprule
	\bf{AU}&\tabincell{c}{Feature \\graph i}& \tabincell{c}{Actural\\AU picture}&\tabincell{c}{Deconvolution\\feature graph}&\tabincell{c}{Origina\\picture}\\
	\midrule
	1&113&\includegraphics[height=0.35in]{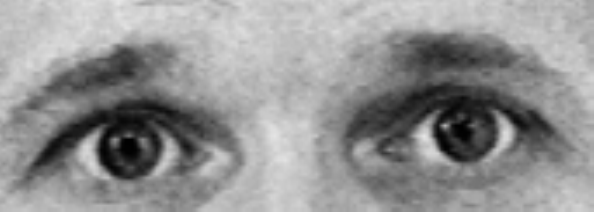}&\includegraphics[height=0.35in]{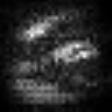}
	&\includegraphics[height=0.35in]{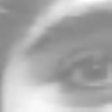}\\
	\bottomrule
\end{tabular}
\end{table}

\begin{table}[h]
\centering
\caption{The corresponding deconvolution feature graph of AU of Figure 6}
\begin{tabular}{p{0.5cm}ccc}
	\toprule
	\bf{AU}&\tabincell{c}{actural \\AU picture}& \tabincell{c}{Decon-\\volution}&\tabincell{c}{Original\\picture}\\
	\midrule
	4&\includegraphics[height=0.35in]{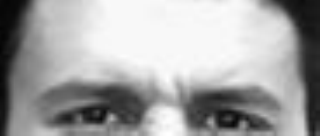}&\includegraphics[height=0.35in]{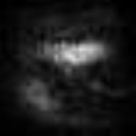}
	&\includegraphics[height=0.35in]{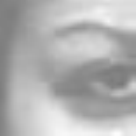}\\
	1&\includegraphics[height=0.35in]{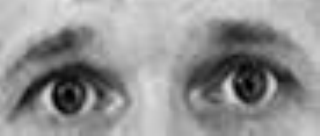}&\includegraphics[height=0.35in]{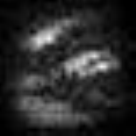}
	&\includegraphics[height=0.35in]{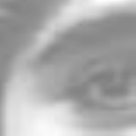}\\
	11&\includegraphics[height=0.35in]{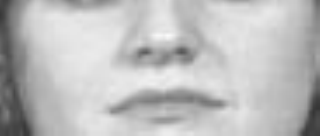}&\includegraphics[height=0.35in]{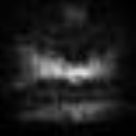}
	&\includegraphics[height=0.35in]{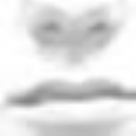}\\
	27&\includegraphics[height=0.35in]{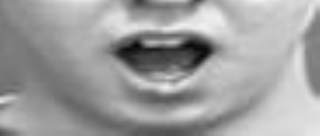}&\includegraphics[height=0.35in]{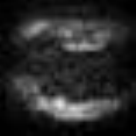}
	&\includegraphics[height=0.35in]{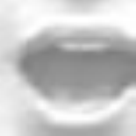}\\
	
	\bottomrule
\end{tabular}
\end{table}

Therefore, the network does learn the facial action unit detector and we verify that in expression classification problem, the principle of convolution neural network is based on the combination of appearance, through the network to detect the specific action unit to classify the face.

In this section, by calculating the response of facial feature units on different feature graphs, the distance function D is designed to measure the relationship between facial feature units and feature graphs and we found the corresponding relationship between facial feature units and feature graphs. The subjective results obtained in section 3 are verified experimentally. It not only confirms that the trained network forms the detector of the specific image, but also obtains the mapping relation between the detector of the network and the designated facial feature unit.

\section{Summary}
In this article, we analyze and verify the principle of expression classification based on a convolution neural network model. The connection between convolution neural network and traditional pattern recognition method by constructing detector is established. And the principle of convolution neural network in classification is verified intuitively: the trained convolution neural network forms the detector of specific image. By calculating the distance function of a facial action unit on different feature graphs, we establish the corresponding relationship between the feature graphs and the action units in the network synchronously, which further supports the hypothesis of the classification principle of the convolution neural network proposed. This conclusion has been verified in the expression classification problem, which can be inferred to be equally applicable in other issues.

\bibliographystyle{IEEEtranS}
\bibliography{D3D-DRL}
	
\end{document}